\documentclass[10pt,twocolumn,letterpaper]{article}

\usepackage{wacv}      

\usepackage{graphicx}
\usepackage{amsmath}
\usepackage{amssymb}
\usepackage{booktabs}

\usepackage{algorithm}
\usepackage{algpseudocode}
\usepackage{amsmath}
\usepackage{booktabs}
\usepackage{siunitx}
\usepackage[numbers,sort&compress]{natbib}

\usepackage[pagebackref,breaklinks,colorlinks]{hyperref}

\usepackage[capitalize]{cleveref}
\crefname{section}{Sec.}{Secs.}
\Crefname{section}{Section}{Sections}
\Crefname{table}{Table}{Tables}
\crefname{table}{Tab.}{Tabs.}

\def\wacvPaperID{*****} 
\def\confName{WACV}
\def\confYear{2025}

\begin{document}

\title{Real-World Transferable Adversarial Attack \\ on Face-Recognition Systems}

\author{
Andrey Kaznacheev$^{5}$,
Matvey Mikhalchuk$^{2}$,\\
Andrey Kuznetsov$^{3,2}$,
Aleksandr Petiushko$^{4}$,
Anton Razzhigaev$^{2,5}$\\
$^2$FusionBrain Lab,
$^3$Innopolis University,
$^4$Elea AI,
$^5$AbstractDL\\
{\tt\small anton.razzhigaev@gmail.com}
}

\maketitle

\begin{abstract}
Adversarial attacks on face recognition (FR) systems pose a significant security threat, yet most are confined to the digital domain or require white-box access. We introduce \textbf{GaP (Gaussian Patch)}, a novel method to generate a universal, physically transferable adversarial patch under a strict black-box setting. Our approach uses a query-efficient, zero-order greedy algorithm to iteratively construct a symmetric, grayscale pattern for the forehead. The patch is optimized by successively adding Gaussian blobs, guided only by the cosine similarity scores from a surrogate FR model to maximally degrade identity recognition. We demonstrate that with approximately 10,000 queries to a black-box ArcFace model, the resulting GaP achieves a high attack success rate in both digital and real-world physical tests. Critically, the attack shows strong transferability, successfully deceiving an entirely unseen FaceNet model. Our work highlights a practical and severe vulnerability, proving that robust, transferable attacks can be crafted with limited knowledge of the target system.

The code will be released after publication.
\end{abstract}

\begin{figure}[h]
\centering
\includegraphics[width=\columnwidth]{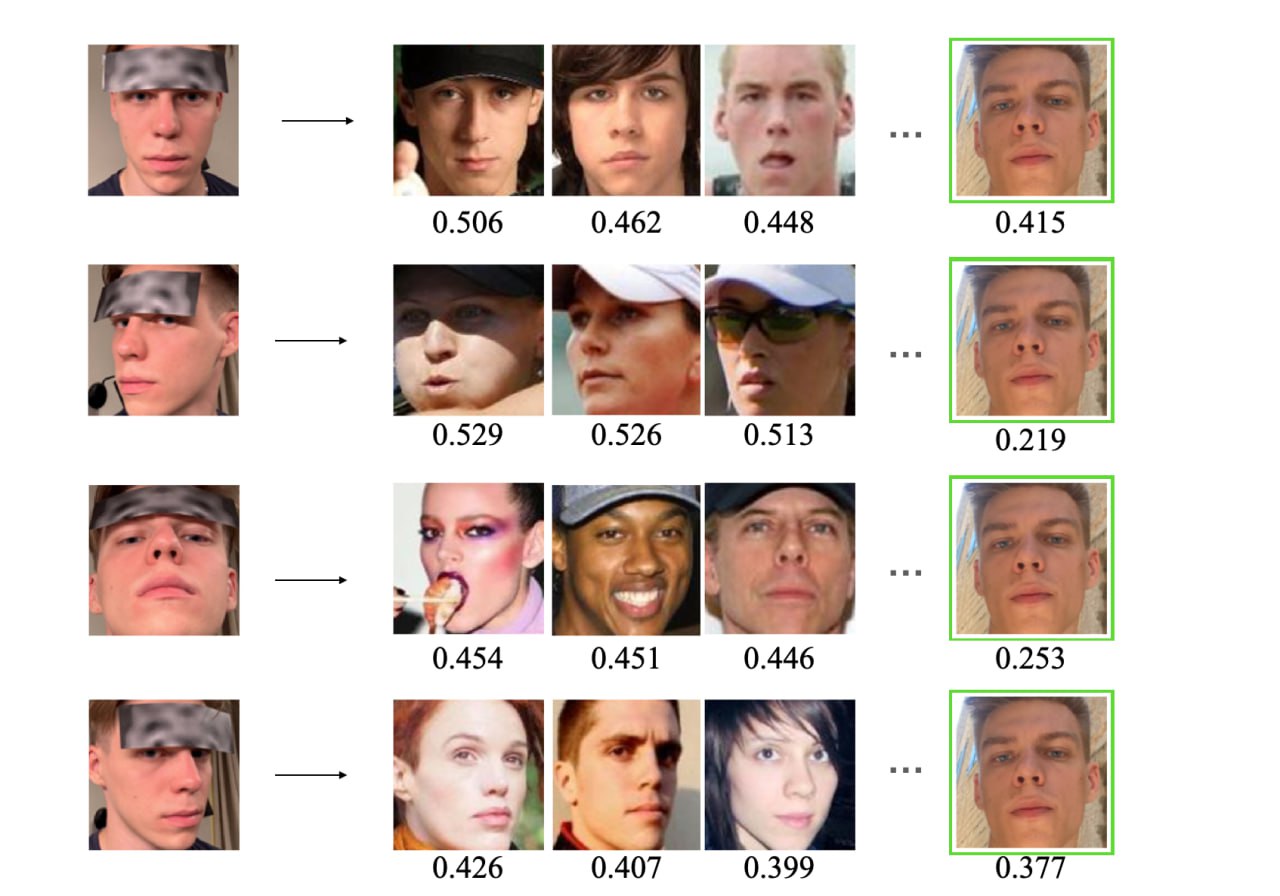}
\caption{Real-world attack examples on video frames. The person wearing the GaP is on the left. The model retrieves the top-3 most similar faces from CelebA (center) and the subject's own most similar reference photo (right, green box). In most cases, the similarity score to unrelated faces is higher than to the correct identity, demonstrating a successful attack.}
\label{fig:examples}
\end{figure}


\section{Introduction}

Deep neural networks have brought face recognition (FR) systems to superhuman accuracy, leading to their widespread deployment in security-critical applications like biometric authentication and surveillance \citep{deng_arcface, schroff_facenet}. Despite their performance, these systems are fundamentally vulnerable to adversarial examples—inputs with carefully crafted perturbations that cause misclassification \citep{szegedy2014intriguing, goodfellow2015explaining}. While initial research focused on digital manipulations, the most tangible threat comes from attacks that are effective in the physical world \citep{kurakin2016adversarial, athalye2018synthesizing}.

A successful physical attack presents a formidable challenge. It must remain effective across a range of uncontrolled real-world conditions, including variable lighting, camera angles, and distances. Furthermore, real-world attackers seldom have internal access to the target system's architecture or weights. This necessitates attacks that operate in a \textbf{black-box} setting, relying only on the model's output scores or decisions \citep{chen2017zoo, brendel2018decision}. The ultimate goal is to create a perturbation that is not only robust and query-efficient but also \textbf{transferable}, enabling it to compromise different, unseen FR models \citep{liu2017transferable}.

In this paper, we address these challenges by proposing a new method for creating a robust and transferable physical adversarial patch. We call our method \textbf{GaP}, for \textbf{Gaussian Patch}, which reflects its core generative process. GaP employs a zero-order, greedy search algorithm to iteratively build a grayscale, symmetric pattern designed to be worn on the forehead. The optimization process starts with a blank slate and progressively adds Gaussian blobs to the patch. The placement and properties of each blob are guided by a simple objective: to achieve the greatest reduction in the identity similarity score returned by a black-box surrogate model. This query-based approach requires no gradients or model knowledge.

We experimentally validate GaP by optimizing a patch using only the output scores from a state-of-the-art ArcFace model. Our results show that the generated patch is highly effective at evading the source model in both digital simulations and physical print-and-wear trials. More importantly, we demonstrate that the GaP exhibits remarkable transferability, successfully deceiving an unseen FaceNet model, thereby exposing a critical and practical vulnerability in modern FR technologies.

Our main contributions are:
\begin{itemize}
    \item We introduce \textbf{GaP}, a novel, query-efficient algorithm for generating physical adversarial patches in a practical black-box setting using a greedy, Gaussian-based search.
    \item We demonstrate that the generated patch is robust and highly effective in real-world physical scenarios against a state-of-the-art face recognition model.
    \item We provide strong evidence of the attack's transferability, showcasing its ability to deceive a different, non-surrogate model, which highlights a serious vulnerability in current FR systems.
\end{itemize}

\section{Related Work}

\subsection{Early Digital Attacks}
The fragility of deep networks was revealed by \citep{szegedy2014intriguing}, who showed that imperceptible $\ell_{\infty}$-noise flips ImageNet predictions.  
\citep{goodfellow2015explaining} explained the effect by "excessive linearity" and proposed FGSM for one-shot generation.

\subsection{Bringing Perturbations into Reality}
\citep{kurakin2016adversarial} proved that printed adversarial images survive a camera pipeline.  
Wearable attacks soon followed: eyeglass frames~\citep{sharif2016accessorize}, expectation-over-transformation 3-D objects~\citep{athalye2018synthesizing}, and sticker-on-hat patches~\citep{komkov2019advhat}.

\subsection{Black-Box Optimisation and Transferability}
Without gradients, attackers rely on zeroth-order queries~\citep{chen2017zoo} or purely decision-based walks~\citep{brendel2018decision}.  
Transfer-based methods~\citep{liu2017transferable} craft examples on a surrogate ensemble that generalise to unseen models.

\subsection{Universal and Wearable Patches}
Universal, person-agnostic artefacts raise the bar even higher.  
A monochrome forehead patch for ArcFace-100 was introduced by \citep{pautov2019arcface_patch}.  
The Adversarial Mask~\citep{zolfi2021adversarial_mask} hides in plain sight on common face masks.  
Generative-manifold regularisation improves cross-model success rates~\citep{xiao2021transferable_patch}, while NatMask~\citep{xie2024natmask} yields highly natural-looking 3-D masks that bypass four SOTA FR models.

Recent universal-wearable advances.
\citep{wei2021adversarialsticker} propose Adversarial Sticker, a physically feasible patch printed on everyday stickers whose pose (position and rotation) is optimised with a region-based heuristic differential-evolution search. In black-box, real-world tests on LFW their method achieves a 76. More recently, \citep{zolfi2021adversarialmask} introduce Adversarial Mask—a fabric face mask imprinted with a universal perturbation that lowers ArcFace’s mean cosine similarity by 0.554 on CelebA and reduces CCTV re-identification to 3.34.

\subsection{Commercial cloud face-recognition APIs} Existing APIs impose tight throughput ceilings that make large-scale query-based attacks costly in time and money: Azure Face limits the free tier to 20 transactions per minute and paid tiers to a default 10 transactions per second (TPS) per resource\footnote{\url{https://learn.microsoft.com/en-us/azure/ai-services/computer-vision/identity-quotas-limits}}; Amazon Rekognition allows only 25–100 TPS for image endpoints such as \texttt{DetectFaces} and just 5 TPS for collection operations like \texttt{AssociateFaces}, depending on the region\footnote{\url{https://docs.aws.amazon.com/general/latest/gr/rekognition.html}}; Face++’s free key shares a global pool and commercial plans start at roughly 3 QPS for face recognition (1 QPS for dense landmarks) unless extra “QPS capacity’’ is purchased\footnote{\url{https://itrexgroup.com/blog/how-much-does-a-facial-recognition-system-cost/}}. These practical caps translate into minutes or hours of wall-clock time when tens of thousands of queries are needed, reinforcing the motivation for query-efficient black-box attacks such as our \emph{GaP} patch.

Existing physical attacks either need heavy query budgets or lose transferability.  
We propose a greedy zero-order optimiser that stays query-efficient, preserves wearable aesthetics, and transfers from ArcFace to FaceNet under a strict black-box assumption.

\section{The GaP Method}
\begin{figure}[h]
\centering
\includegraphics[width=0.9\columnwidth]{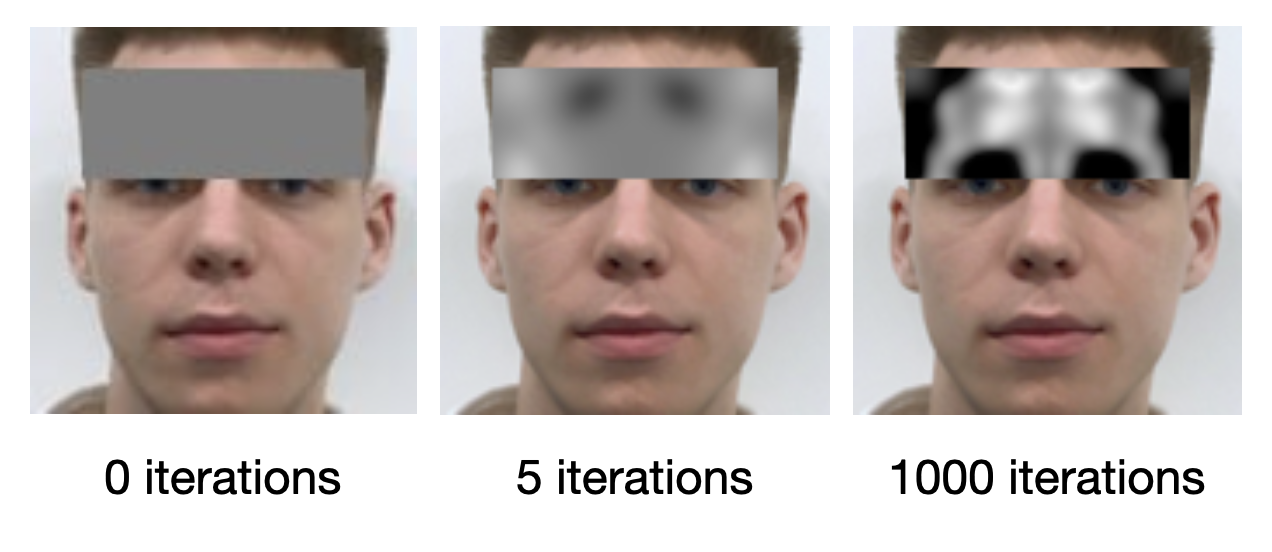}
\caption{Visualization of the intermediate iterations of \textbf{GaP} algorithms gradually adding Gaussian blobs on the attacked image.}
\label{fig:progress}
\end{figure}

We propose \textbf{GaP (Gaussian Patch)}, a method to craft a physical adversarial patch in a strict black-box setting. The core of our approach is a query-efficient, zero-order greedy algorithm that builds the patch by iteratively adding Gaussian blobs to minimize an FR system's similarity score (see Figure~\ref{fig:progress})

\subsection{Threat Model and Attack Goal}
Our threat model assumes a practical \textbf{black-box} setting. The attacker has no knowledge of the target Face Recognition (FR) model's architecture or parameters. The only capability is to perform queries: submitting two images to receive their cosine similarity score.

The goal is a universal \textbf{evasion attack} (also known as dodging). We aim to create a single patch that, when applied to a person's face, significantly lowers the similarity score between the armed image and a clean reference image of the same individual, causing the system to fail verification.

\subsection{Patch Design and Application}
To ensure physical feasibility and create a structured appearance, the GaP is designed with the following constraints:
\begin{itemize}
    \item \textbf{Form:} The patch is a rectangular, grayscale pattern designed to be printed and worn on the forehead, a common location for wearable sensors or decorative elements.
    \item \textbf{Symmetry:} We enforce bilateral symmetry to create a more natural pattern and reduce the optimization search space. The algorithm optimizes only one half of the patch, $P_{\text{left}}$, and mirrors it to create the other half: $P_{\text{right}} = \text{flip}(P_{\text{left}})$.
    \item \textbf{Application:} In the digital domain, the patch $P$ is overlaid at a fixed position onto face images that have been aligned and resized to $112 \times 112$ pixels using MTCNN.
\end{itemize}

\subsection{Greedy Optimization with Gaussian Blobs}
Lacking gradient information, we employ a \textbf{zero-order}, greedy search algorithm. The process begins with a blank (zero) patch and iteratively refines it. Each iteration involves four key steps:

\begin{enumerate}
    \item \textbf{Candidate Generation:} We generate a batch of $B$ random elliptical Gaussian blobs $\{g_1, \dots, g_B\}$. Each blob $g_i$ is a potential addition to the patch, defined by a randomly sampled amplitude, center, variance, and rotation.

    \item \textbf{Query-Based Evaluation:} Each blob is added to the current best patch, $P_{t-1}$, to form a candidate patch $P_{t,i}$. We evaluate each candidate by querying the surrogate model (ArcFace) using a loss function $\mathcal{L}$. To improve robustness, the loss minimizes the similarity score over two different source images, $I_A$ and $I_B$, of the same identity:
    \begin{equation}
    \mathcal{L}(P) = \sum_{x, y \in \{A, B\}} \text{sim}(\text{Apply}(I_x, P), I_y)
    \label{eq:loss}
    \end{equation}
    where $\text{sim}$ is the FR model's similarity response and $\text{Apply}(\cdot, \cdot)$ overlays the patch on an image.

    \item \textbf{Greedy Selection:} The candidate patch that yields the minimum loss, $P_t = \arg\min_{P_{t,i}} \mathcal{L}(P_{t,i})$, is selected and becomes the base for the next iteration.

    \item \textbf{Restart Mechanism:} To mitigate convergence to poor local minima, we integrate a \textbf{restart} strategy. The optimization is reset to a blank patch every $N_{\text{restart}}$ iterations. A separate variable tracks the globally best patch found across all iterations and restarts, ensuring progress is never lost.
\end{enumerate}

This iterative search, formalized in Algorithm \ref{alg:patch_generation}, is run for a fixed budget of queries to produce the final adversarial patch.

\begin{figure}[t]
\centering
\includegraphics[width=0.9\columnwidth]{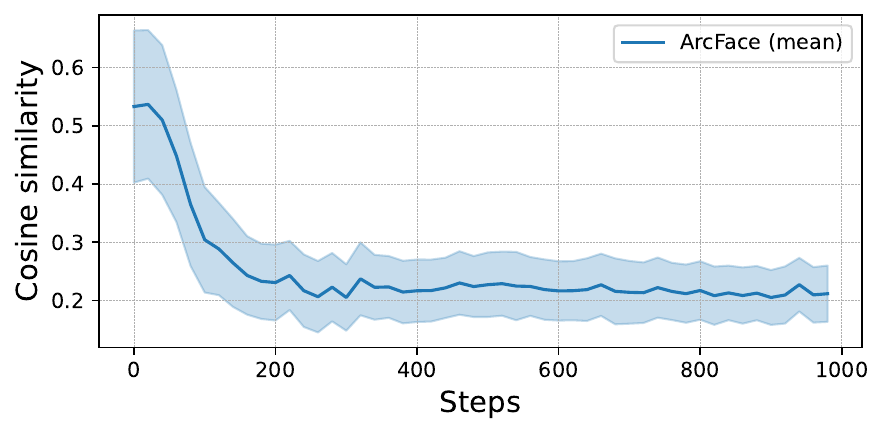}
\caption{Cosine similarity score of the attacked image as a function of optimization iterations. Each iteration comprises 32 queries to the black-box model. The score consistently decreases, showing the attack's effectiveness over a budget of approximately 20,000 queries.}
\label{fig:scores}
\end{figure}

\begin{algorithm}[h]
\caption{Greedy Gaussian-based Adversarial Patch Generation}
\label{alg:patch_generation}
\begin{algorithmic}[1] 
\State \textbf{Input:} Source images $I_A, I_B$ of the same identity.
\State \textbf{Input:} Iterations $N_{iters}$, batch size $B$, restart interval $N_{restart}$.
\State \textbf{Initialize:} Current patch $P_{curr} \leftarrow \mathbf{0}$.
\State \textbf{Initialize:} Global best patch $P_{best} \leftarrow \mathbf{0}$.
\State \textbf{Initialize:} Global best score $\mathcal{L}_{best} \leftarrow \infty$.

\For{$t = 1 \to N_{iters}$}
    \State Generate a batch of $B$ random Gaussian blobs $\{g_1, \dots, g_B\}$.
    \State $\mathcal{L}_{batch\_best} \leftarrow \infty$.
    \State $P_{batch\_best} \leftarrow P_{curr}$.

    \For{$i = 1 \to B$}
        \State $P_{cand} \leftarrow \text{clamp}(P_{curr} + g_i, -1, 1)$.
        \State $\mathcal{L}_{cand} \leftarrow \text{CalculateLoss}(P_{cand}, I_A, I_B)$ using Eq. \ref{eq:loss}.
        \If{$\mathcal{L}_{cand} < \mathcal{L}_{batch\_best}$}
            \State $\mathcal{L}_{batch\_best} \leftarrow \mathcal{L}_{cand}$.
            \State $P_{batch\_best} \leftarrow P_{cand}$.
        \EndIf
    \EndFor
    
    \State $P_{curr} \leftarrow P_{batch\_best}$ \Comment{Greedy update}
    
    \If{$\mathcal{L}_{batch\_best} < \mathcal{L}_{best}$}
        \State $\mathcal{L}_{best} \leftarrow \mathcal{L}_{batch\_best}$.
        \State $P_{best} \leftarrow P_{curr}$. \Comment{Update global best}
    \EndIf
    
    \If{$t \pmod{N_{restart}} = 0$ and $t < N_{iters}$}
        \State $P_{curr} \leftarrow \mathbf{0}$ \Comment{Restart search}
    \EndIf
\EndFor
\State \textbf{Return} $P_{best}$
\end{algorithmic}
\end{algorithm}

Figure~\ref{fig:scores} illustrates the optimization process, plotting the cosine similarity of the attacked image against the number of iterations. Each iteration corresponds to approximately 32 queries, and the plot shows a consistent decrease in the similarity score as the algorithm progresses. This demonstrates the query-efficiency of our greedy search in finding a potent adversarial solution within a budget of roughly 20,000 total queries.

\section{Experiments}
\label{sec:experiments}

\begin{table*}
\centering
\caption{Attack Success Rate (\%) on LFW and CelebA. We add a third LFW column for \textbf{SphereFace}. `--` = not reported.}
\label{tab:comparison}
\begin{tabular}{@{}lccccc@{}}
\toprule
& \multicolumn{3}{c}{\textbf{LFW}} & \multicolumn{2}{c}{\textbf{CelebA}} \\
\cmidrule(lr){2-4} \cmidrule(lr){5-6}
\textbf{Method} & \textbf{ArcFace} & \textbf{FaceNet} & \textbf{SphereFace} & \textbf{ArcFace} & \textbf{FaceNet} \\
\midrule
\multicolumn{6}{@{}l}{\textit{Prior Wearable Attacks}} \\
Adv. Sticker \citep{wei2021adversarialsticker} & -- & 76.3 & 72.93 & -- & 81.8 \\
\addlinespace[3pt]
\multicolumn{6}{@{}l}{\textit{Our Baselines}} \\
Gray Rectangle & 2.3 & 15.9 & 29.4 & 2.4 & 23.8 \\
Forehead Graft & 5.7 & 68.2 & 58.1 & 6.0 & 64.3 \\
\addlinespace[3pt]
\multicolumn{6}{@{}l}{\textit{Our Method}} \\
\textbf{GaP (ours)} & \textbf{80.7} & \textbf{86.4} & \textbf{98.9} & \textbf{69.0} & \textbf{86.9} \\
\bottomrule
\end{tabular}
\end{table*}

\begin{table}[h]
\centering
\caption{Physical-world attack performance. Frame-level misidentification rate (\%) for 10 subjects under varied lighting conditions. The GaP was optimized on ArcFace (source model) and tested for transferability on FaceNet.}
\label{tab:physical_results}
\sisetup{separate-uncertainty, table-format=2.1(1)}
\begin{tabular}{@{}l S S@{}}
\toprule
\textbf{Condition} & \multicolumn{2}{c}{\textbf{Misidentification Rate (\%)}} \\
\cmidrule(l){2-3}
 & {\textbf{ArcFace (Source)}} & {\textbf{FaceNet (Transfer)}} \\
\midrule
\addlinespace[2pt]
\multicolumn{3}{@{}l}{\textbf{Verification against LFW Database}} \\
\cmidrule(r){1-1}
Indoor Lighting    & 63.4(2.1) & 92.0 (2.7) \\
Outdoor Lighting   & 65.5 (2.0) & 96.0 (3.2) \\
\addlinespace[5pt]
\multicolumn{3}{@{}l}{\textbf{Verification against CelebA Database}} \\
\cmidrule(r){1-1}
Indoor Lighting    & 66.6 (4.3) & 99.1 (1.2) \\
Outdoor Lighting   & 71.6 (3.6) & 99.4 (1.2) \\
\bottomrule
\end{tabular}
\end{table}

\begin{table*}[h]
\centering
\caption{Ablation of design choices. Each row is a different configuration of patch constraints and optimizer. We report Attack Success Rate (ASR, \%) on digital LFW. \textbf{Bold} indicates the default/final configuration.}
\label{tab:ablations_combinations}
\begin{tabular}{@{}lccc|cc@{}}

\toprule
\textbf{Symmetry} & \textbf{Color} & \textbf{Restarts} & \textbf{ArcFace} & \textbf{FaceNet} \\
\midrule
\textbf{Yes}   & \textbf{Grayscale} & \textbf{Yes}    & \textbf{80.7} & \textbf{86.4} \\
Yes            & Color        & Yes                     & 70.1           & 58.2 \\
No             & Grayscale    & Yes                     & 65.4           & 68.5 \\
Yes   & Grayscale & No                      & 58.0           & 68.7 \\
\bottomrule
\end{tabular}
\end{table*}

To validate GaP, we first establish our setup, then analyze the patch's geometry, and finally, benchmark it against prior work and new baselines in digital and physical tests.

\subsection{Experimental Setup}
\paragraph{Models and Datasets.} We use \textbf{ArcFace} (ResNet-100) \citep{deng_arcface} as our black-box surrogate model and test transferability on an unseen \textbf{FaceNet} \citep{schroff_facenet}. Experiments are run on the Labeled Faces in the Wild (LFW) \citep{LFWTech} and CelebA \citep{liu2015faceattributes} datasets. Faces are aligned via MTCNN to $112 \times 112$ pixels.

\paragraph{Metrics.} We measure the \textbf{Attack Success Rate (ASR)}, which is the percentage of pairs whose similarity drops below the model's verification threshold. A higher ASR indicates a more effective attack.

\begin{figure}[h]
\centering
\includegraphics[width=1.0\columnwidth]{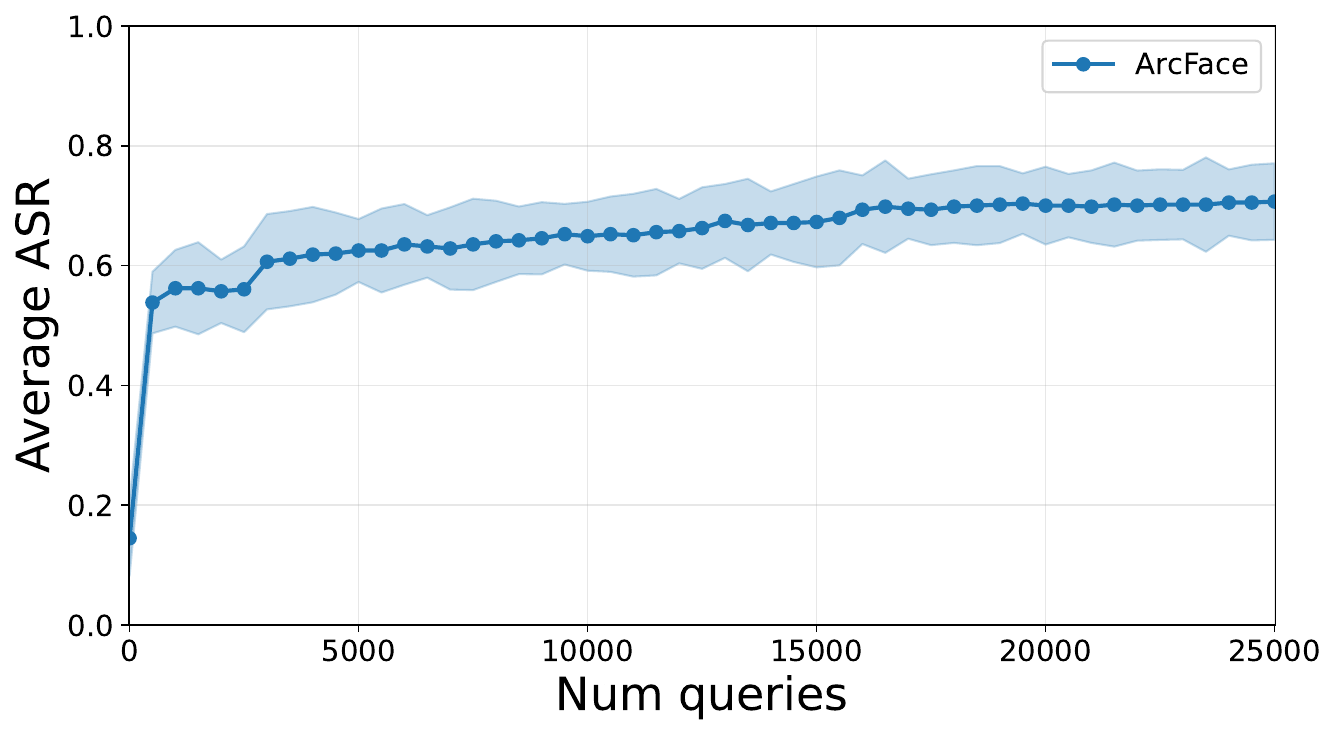}
\caption{Attack Success Rate (ASR) as a function of the number of queries to the ArcFace model. Results are averaged over 5 independent runs on random subsets of 100 identities from LFW.}
\label{fig:num_queries_asr}
\end{figure}

\paragraph{Query Efficiency and Convergence.}
To further demonstrate the efficiency of our query-based optimization, we analyze the attack success rate (ASR) as a function of the number of queries used during the patch generation process. Figure~\ref{fig:num_queries_asr} reports the ASR averaged over 5 independent runs, each on a random subset of 100 identities from the LFW dataset, attacking ArcFace. The results show that the proposed algorithm achieves rapid convergence: a substantial portion of the final ASR is already reached within the first few thousand queries. This indicates that the patch optimization process is not only effective, but also highly query-efficient, making it practical even under the tight query-rate constraints imposed by real-world commercial APIs.

\section{Ablation Study}
\label{sec:ablations}

To better understand which design decisions are most critical to GaP's effectiveness and transferability, we conduct a systematic ablation study. Specifically, we isolate and evaluate the contributions of three key factors: (1) bilateral symmetry, (2) color vs. grayscale, and (3) use of optimization restarts. All ablations are performed on the digital LFW setup using ArcFace as the source model; for each configuration, we report Attack Success Rate (ASR, \%).

\subsection{Effect of Symmetry, Color, and Restarts}

Table~\ref{tab:ablations_combinations} presents the results for all meaningful combinations of these factors. Each experiment modifies a single aspect of the default setup (symmetric, grayscale, with restarts) to assess its impact.

\paragraph{Symmetry.}
Removing the bilateral symmetry constraint leads to a significant drop in attack success (from 80.7\% to 65.4\%). This confirms that constraining the patch to be symmetric both reduces the search space and, perhaps surprisingly, produces more transferable and robust adversarial patterns. Non-symmetric patterns are less effective and less physically plausible.

\paragraph{Color.}
Allowing full color in the patch (instead of restricting to grayscale) also reduces performance, with ASR falling to 70.1\%. We hypothesize that high-frequency chromatic variation is less effective at manipulating deep face features, or is suppressed by face alignment and normalization pipelines in FR systems. Thus, grayscale constraints both simplify optimization and yield stronger attacks.

\paragraph{Restarts.}
Disabling the optimizer's restart mechanism causes the attack to degrade sharply to 58.0\% ASR, highlighting the importance of escaping poor local minima. Restarts consistently improved final performance in all trials.

\paragraph{Summary.}
Combining all ablations, our default setup (symmetric, grayscale, with restarts) provides the strongest attack, and is adopted throughout the rest of the paper. For FaceNet, we observe a similar trend: every deviation from the default setup reduces transfer attack performance (detailed numbers are in preparation).

\subsection{Ablation of Patch Geometry}
\label{sec:ablations_geometry}

\begin{figure}[h]
\centering
\includegraphics[width=0.9\columnwidth]{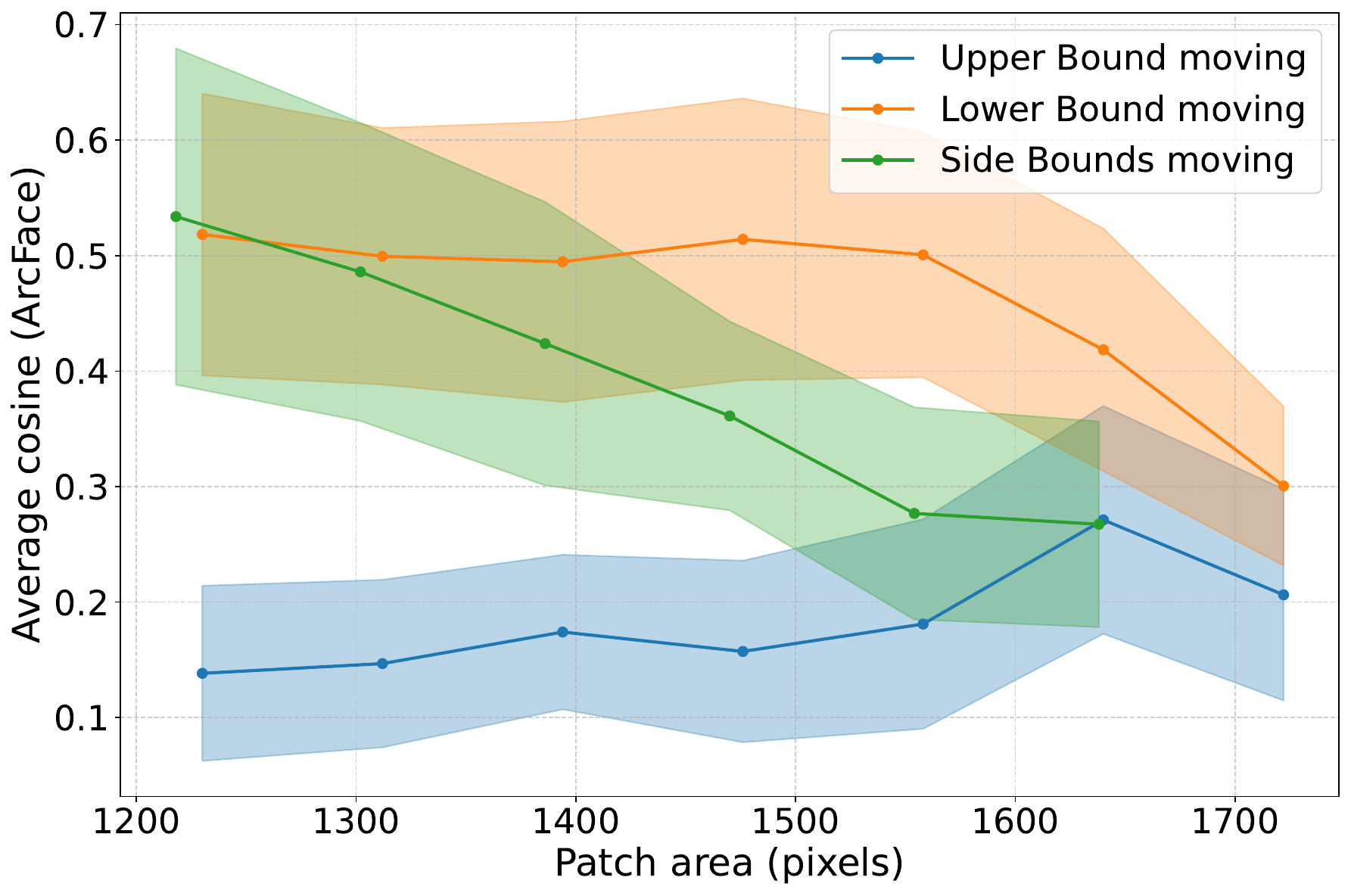}
\caption{Effect of patch geometry. The attack’s effectiveness is highly localized: removing the lower edge (near the eyebrows) almost eliminates the effect, while trimming the top has little impact. A narrow lower-central band retains most of the attack’s power.}
\label{fig:patch_geometry}
\end{figure}

We further investigate the effect of patch shape and position. Figure~\ref{fig:patch_geometry} shows the impact of shrinking the patch either from the top (toward the hairline) or the bottom (toward the eyebrows), and of reducing it to a narrow central band. We measure the average cosine similarity (lower is better).

\paragraph{Findings.}
The lower-central region of the patch (just above the eyebrows) is critical for effectiveness; this may reflect the high concentration of identity features in this facial region. The top part of the patch is redundant and can be removed with little loss in performance, further supporting the practicality and visual unobtrusiveness of the attack.

\subsection{Comparative and Physical Evaluation}
\begin{figure}[h]
\centering
\includegraphics[width=0.9\columnwidth]{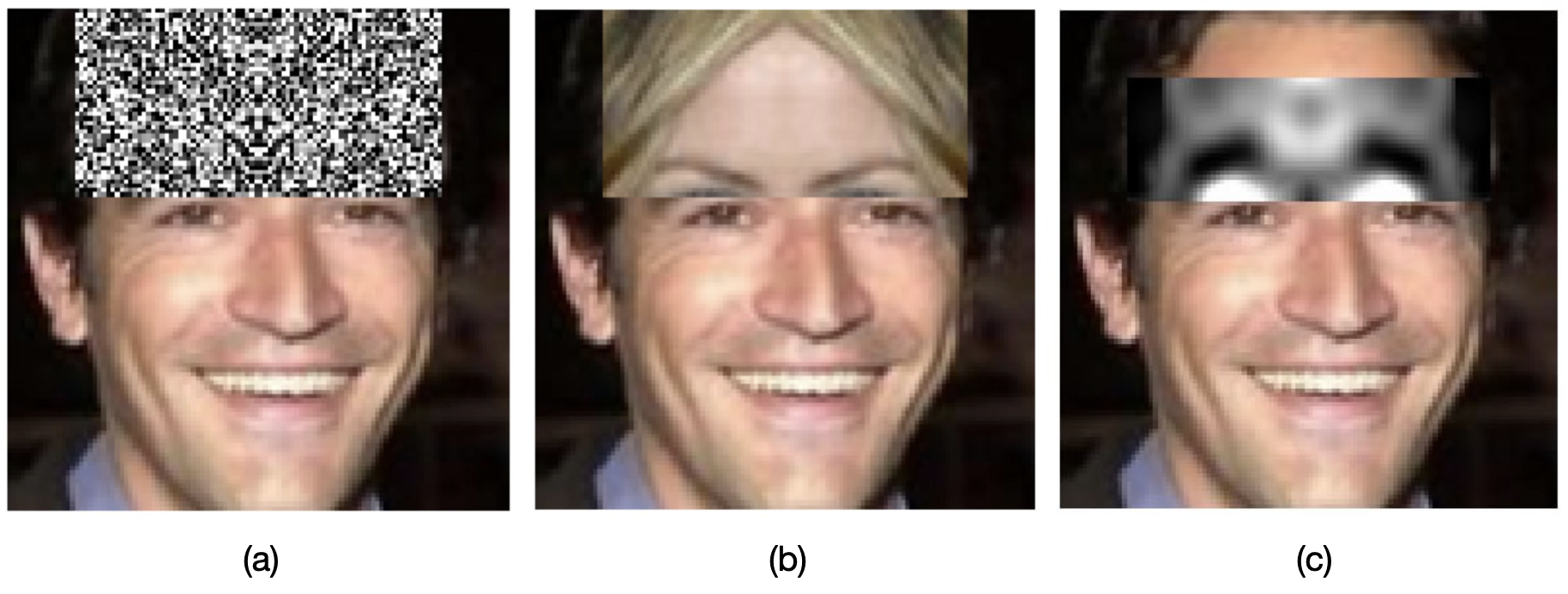}
\caption{Patches used in the evaluation: (a) Random noise patch , (b) Forehead Graft baseline, and (c) Our optimized GaP.}
\label{fig:baselines}
\end{figure}

We benchmark GaP against other state-of-the-art physical attacks and two strong baselines, shown in Figure~\ref{fig:baselines}. The first baseline is a simple \textbf{Noise Rectangle}, which tests the effect of plain occlusion. The second is a \textbf{Forehead Graft}, where a forehead patch from an unrelated identity is used, testing a naive but texture-rich adversarial prior.

Table~\ref{tab:comparison} summarizes the performance. The Noise Rectangle has a minimal effect on the source model (ArcFace) but shows some impact on FaceNet, suggesting the latter is more sensitive to simple occlusion. The Forehead Graft proves to be a formidable baseline, especially in transfer attacks, achieving over 64\% ASR on FaceNet. 

Our method, \textbf{GaP}, significantly outperforms these baselines. It achieves a high ASR against the source model (80.7\% on LFW, 69.0\% on CelebA), demonstrating the efficacy of our query-based optimization. Critically, GaP also shows superior transferability, reaching over 86\% ASR on the unseen FaceNet model for both datasets, surpassing all other methods.

\paragraph{Physical-World Evaluation.}
To rigorously validate the patch's efficacy in the physical domain, we conducted an experiment with 10 participants. Each subject was recorded in 60-second videos under two conditions: indoor room lighting and outdoor daylight. During recordings, participants walked toward the camera while making natural head movements within a 15-degree range. We measured the frame-level misidentification rate against both LFW and CelebA databases, which were augmented with 10 reference images of each participant. As detailed in Table \ref{tab:physical_results}, the patch demonstrated robust real-world performance, achieving high misidentification rates across all scenarios. Crucially, the attack maintained strong transferability to the unseen FaceNet model.

Figure \ref{fig:examples} provides a qualitative illustration of this real-world impact. In several video frames, the patch successfully deceives the model, causing it to report a higher cosine similarity to unrelated identities from the CelebA database than to the subject’s own reference photo.

\section{Conclusion}

In this work, we introduced \textbf{GaP}, a novel method to generate a universal, physically transferable adversarial patch for face recognition systems under a strict black-box constraint. Our query-efficient, greedy algorithm constructs the patch by iteratively adding Gaussian blobs, guided solely by the similarity scores of a surrogate model. We demonstrated that the resulting patch is highly effective in both digital and real-world physical tests against the source model, ArcFace. Critically, we showed that the attack exhibits strong transferability, successfully deceiving an entirely unseen FaceNet model with a high success rate. Our findings highlight a practical and severe vulnerability, proving that robust, transferable physical attacks can be crafted with limited knowledge of the target system, posing a tangible threat to deployed FR technologies.

\section{Limitations}
Our study has several limitations. First, the physical-world evaluation was a proof-of-concept conducted with just 10 people under controlled indoor conditions. The patch's robustness to diverse environmental factors (e.g., outdoor lighting, varied camera angles) and demographic variations requires more extensive testing. Second, the attack's transferability was validated only between two prominent CNN-based models (ArcFace and FaceNet). Its effectiveness against commercial, proprietary systems or models with fundamentally different architectures (e.g., Vision Transformers) remains an open question. Finally, the patch is visually conspicuous and would likely be detected by human observers in many real-world security scenarios, limiting its practical stealth. The effect of minor misalignments or partial occlusions on the patch's performance was also not systematically studied.

\section{Ethical Statement}
The research presented in this paper demonstrates a method for creating physical adversarial patches that can evade face recognition systems. We acknowledge the potential for misuse of such technology. However, our work is intended for a defensive purpose: to proactively identify and expose a critical, transferable vulnerability in existing systems under a realistic black-box threat model. By understanding the mechanisms of such attacks, the research community can develop more effective countermeasures. All experiments were conducted in a controlled environment using public datasets and with the full consent of the participants involved in the physical evaluation.

{\small
\bibliographystyle{ieee_fullname}
\bibliography{aaai2026}
}

\end{document}